\documentclass{article}

\usepackage{amsfonts}
\usepackage{amsmath}
\usepackage{booktabs}
\usepackage{color}
\usepackage{floatrow}
\usepackage{graphicx}
\usepackage{hhline}
\usepackage{hyperref}
\usepackage{microtype}
\usepackage{multirow}
\usepackage{subfigure}

\usepackage[accepted]{icml2021}

\icmltitlerunning{Nondeterminism and Instability in Neural Network Optimization}

\begin{document}

\twocolumn[
\icmltitle{Nondeterminism and Instability in Neural Network Optimization}

\icmlsetsymbol{equal}{*}

\begin{icmlauthorlist}
\icmlauthor{Cecilia Summers}{auck}
\icmlauthor{Michael J. Dinneen}{auck}
\end{icmlauthorlist}

\icmlaffiliation{auck}{Department of Computer Science, University of Auckland, Auckland, New Zealand}

\icmlcorrespondingauthor{Cecilia Summers}{cecilia.summers.07@gmail.com}

\icmlkeywords{Nondeterminism, Instability, Optimization}

\vskip 0.3in
]

\printAffiliationsAndNotice{}

\begin{abstract}
Nondeterminism in neural network optimization produces uncertainty in performance, making small improvements difficult to discern from run-to-run variability.
While uncertainty can be reduced by training multiple model copies, doing so is time-consuming, costly, and harms reproducibility.
In this work, we establish an experimental protocol for understanding the effect of optimization nondeterminism on model diversity, allowing us to isolate the effects of a variety of sources of nondeterminism.
Surprisingly, we find that all sources of nondeterminism have similar effects on measures of model diversity.
To explain this intriguing fact, we identify the instability of model training, taken as an end-to-end procedure, as the key determinant.
We show that even one-bit changes in initial parameters result in models converging to vastly different values.
Last, we propose two approaches for reducing the effects of instability on run-to-run variability.
\end{abstract}

\section{Introduction}
\label{sec:introduction}
Consider this common scenario: you have a baseline ``current best'' model, and are trying to improve it.
One of your experiments has produced a model whose metrics are slightly better than the baseline.
Yet you have your reservations --- how do you know the improvement is ``real'' and not due to run-to-run variability?

Similarly, consider hyperparameter optimization, in which many possible values exist for a set of hyperparameters, with minor differences in performance between them. How do you pick the best hyperparameters, and how can you be sure that you've actually picked wisely?

In both scenarios, the standard practice is to train multiple independent copies of your model to understand its variability.
While this helps address the problem, it is extremely wasteful, using more computing power, increasing the time required for effective research, and making reproducibility difficult, all while still leaving some uncertainty.

Ultimately, the source of this problem is nondeterminism in model optimization --- randomized components of model training that cause each run to produce different models with their own performance characteristics.
Nondeterminism itself occurs due to many factors: while the most salient source is the random initialization of parameters, other sources exist, including random shuffling of training data, stochasticity in data augmentation, explicit random operations (e.g. dropout~\cite{srivastava2014dropout}), asynchronous training~\cite{recht2011hogwild}, and even nondeterminism in low-level libraries such as cuDNN~\cite{chetlur2014cudnn}.

Despite the clear impact nondeterminism has on the efficacy of modeling, relatively little attention has been paid towards understanding its mechanisms.
In this work, we establish an experimental protocol for analyzing the impact of nondeterminism in model training, allowing us to quantify the independent effect of each source of nondeterminism.
In doing so, we make a surprising discovery: each source has nearly the same effect on the variability of final model performance.
Further, we find each source produces models of similar diversity, as measured by correlations between model predictions, functional changes in model performance while ensembling, and state-of-the-art methods of model similarity~\cite{kornblith2019similarity}.
To emphasize one particularly interesting result: nondeterminism in low-level libraries like cuDNN can matter just as much with respect to model diversity and variability as varying the entire network initialization.

We explain this mystery by demonstrating that it can be attributed to \emph{instability} in optimizing neural networks --- 
when training with SGD-like approaches, we show that small changes to initial parameters result in large changes to final parameter values.
In fact, the instabilities in the optimization process are extreme: \emph{changing the initialization of a single weight by the smallest possible amount within machine precision (${\sim}6 \cdot 10^{-11}$) produces nearly as much variability as all other sources combined}. Therefore, any source of nondeterminism with any effect at all on model weights inherits at least this level of variability.

Last, we present promising results in reducing the effects of instability on run-to-run variability.
While we find that many approaches result in no apparent change, we propose and demonstrate two approaches that reduce model variability without any increase in model training time: accelerated model ensembling and test-time augmentation.
Together, these provide the first encouraging signs for the tractability of this problem. Code has been made publicly available.\footnote{\url{https://github.com/ceciliaresearch/nondeterminism_instability}}

\section{Related Work}
\label{sec:related_work}
\subparagraph{Nondeterminism.}
Relatively little prior work has studied the effects of nondeterminism on model optimization.
While nondeterminism is recognized as a significant barrier to reproducibility and evaluating progress in some subfields of machine learning, such as reinforcement learning~\cite{nagarajan2018impact,henderson2018deep,islam2017reproducibility,machado2018revisiting}, in the setting of supervised learning, the focus of this work, the problem is much less studied.
Madhyastha and Jain~\cite{madhyastha2019model} aggregate all sources of nondeterminism together into a single random seed and analyze the variability of model attention and accuracy across various NLP datasets. They also propose a method for reducing this variability (see Appendix Sec.~\ref{sec:didnt_work} for details of our reproduction attempt).
More common in the field, results across multiple random seeds are reported~\cite{erhan2010does}, but the precise nature of nondeterminism's influence on variability goes unstudied.

\subparagraph{Instability.}
We use the term ``stability'' in a manner analogous to numerical stability~\cite{higham2002accuracy}, where a stable algorithm is one for which the final output (converged model) does not vary much as the input (initial parameters) are changed.
In other contexts, the term ``stability'' has been used both in learning theory \cite{bousquet2002stability} and in reference to vanishing and exploding gradients~\cite{haber2017stable}.

\section{Nondeterminism}
\label{sec:nondeterminism}
Many sources of nondeterminism exist in neural network optimization, each of which affects the variability of trained models.
We begin with a very brief overview:

\subparagraph{Parameter Initialization.} When training a model, parameters without preset values are initialized randomly according to a given distribution, \emph{e.g.} a zero-mean Gaussian with variance determined by the number of input connections to the layer~\cite{glorot2010understanding,he2015delving}.

\subparagraph{Data Shuffling.} In stochastic gradient descent, the gradient is approximated on a random subset of examples, commonly implemented by using small batches of data iteratively in a shuffled training dataset~\cite{bottou2012stochastic}.
Shuffling may happen either once, before training, or in between each epoch of training, the variant we use in this work.

\subparagraph{Data Augmentation.} A common practice, data augmentation refers to randomly altering each training example to artificially expand the training dataset~\cite{shorten2019survey}.
For example, randomly flipping images encourages invariance to left/right orientation.

\subparagraph{Stochastic Regularization.}
Some types of regularization, such as Dropout~\cite{srivastava2014dropout}, take the form of stochastic operations internal to a model during training.
Other instances of this include DropConnect~\cite{wan2013regularization} and variable length backpropagation through time~\cite{merity2017regularizing}, among many others.

\subparagraph{Low-level Operations.}
Often underlooked, many libraries that deep learning frameworks are built on, such as cuDNN~\cite{chetlur2014cudnn}, typically run nondeterministically in order to increase the speed of their operations.
This nondeterminism is small when evaluated in the context of a single operation --- in one test we performed it caused an output difference of $0.003\%$.
In the case of cuDNN, the library we test, it is possible to disable nondeterministic behavior at a speed penalty on the order of ${\sim}15\%$.
However, unlike other nondeterminism sources, it is not possible to ``seed'' this; it is only possible to turn it on or off.

\subsection{Protocol for Testing Effects of Nondeterminism}
\label{sec:protocol}
\subparagraph{Performance Variability.}
Our protocol for testing the effects of sources of nondeterminism is based on properly controlling for each source.
Formally, suppose there are $N$ sources of nondeterminism, with source $i$ controlled by seed $S_i$.
To test the effect of source $i$, we keep all values $\{S_j\}_{j \neq i}$ set to a constant, and vary $S_i$ with $R$ different values, where $R$ is the number of independent training runs performed. For sources of nondeterminism which cannot be effectively seeded, such as cuDNN, we indicate one of these values as the deterministic value, which it must be set to when varying the other sources of nondeterminism.

For example, denote $S_1$ the seed for random parameter initialization, $S_2$ for training data shuffling, and $S_3$ for cuDNN, where $S_3 = 1$ is the deterministic value for cuDNN.
To test the effect of random parameter initialization, with a budget of $R = 100$ training runs, we set $S_3$ to the deterministic value of $1$, $S_2$ to an arbitrary constant (typically 1 for simplicity), and test 100 different values of $S_1$. All together, this corresponds to training models for each of $(S_1, S_2, S_3) \in \{(i, 1, 1)\}_{i=1}^{100}$.
To measure variability of a particular evaluation metric (\emph{e.g.} cross-entropy or accuracy for classification), we calculate the standard deviation (across all $R = 100$ models) of the metric.
Note that it is also possible to test the effect of several sources of nondeterminism in tandem this way, \emph{e.g.} by considering $(S_1, S_2, S_3) \in \{(i, i, 0)\}_{i=1}^{R}$ to measure the joint effect of all three sources in this example.

\subparagraph{Representation Diversity.}
We also examine differences in the \emph{representation} of trained models, complementary to variability in test set performance ---
this allows us to differentiate cases where two sources of nondeterminism have similar performance variability but actually produce models with disparate amounts of representational similarity.
In order to rigorously examine this, we consider four distinct analyses of the functional behavior of models:

The first and simplest metric we consider is the average disagreement between pairs of models, with higher disagreement corresponding to higher diversity and variability.
In contrast to our other metrics, this considers only the argmax of a model's predictions, which makes it the most limited but also the most interpretable of the group.
This metric has also been used recently to compare similarity in the context of network ensembles~\cite{fort2019deep}.

Second, we consider the average correlation between the predictions of two models, \emph{i.e.} the expectation (across pairs of models from the same nondeterminism source) of the correlation of predictions, calculated across examples and classes.
Concretely, for a classification task, the predicted logits from each of $R$ models are flattened into vectors of length $N * C$ (with $N$ test examples and $C$ classes), and we calculate the mean correlation coefficient of the predictions across all $\binom{R}{2}$ pairs of models.
We use Spearman's $\rho$ for the correlation coefficient, but note that other metrics are possible and yield similar conclusions.
For this metric, a lower score indicates a more diverse set of models.

The third analysis we perform examines the change in performance in ensembling two models from the same source of nondeterminism.
Intuitively, if a pair of models are completely redundant, then ensembling them would result in no change in performance.
However, if models actually learn different representations, then ensembling should create an improvement, with a greater improvement the greater the diversity in a set of models.
Denoting by $f(S_i)$ some particular evaluation metric $f$ calculated on the predictions of model $S_i$, and $\frac{S_i + S_j}{2}$ the ensemble of models $S_i$ and $S_j$, this metric is formally determined by:
\begin{equation}
  \frac{1}{\binom{R}{2}} \sum_{i=1}^R \sum_{j=i+1}^R \left( f\left(\frac{S_i + S_j}{2}\right) - \frac{f(S_i) + f(S_j)}{2} \right)
\end{equation}
Last, for a more detailed view of learned representations internal to a network, we consider a state-of-the-art method for measuring the similarity of neural network representations, centered kernel alignment (CKA)~\cite{kornblith2019similarity}, which has previously been used to analyze models trained with different random initializations, widths, and even entirely different architectures.
We use the linear version of CKA, which Kornblith \emph{et al.} found to perform similarly to more complicated RBF kernels.

\subsection{Experiments in Image Classification}
We begin our study of nondeterminism with the fundamental task of image classification.
We execute our protocol with CIFAR-10~\cite{krizhevsky2009learning} as a testbed, a 10-way classification dataset with $50{,}000$ training images of resolution $32 \times 32$ pixels and $10{,}000$ images for testing.
In these initial experiments, we use a 14-layer ResNet model~\cite{he2016deep}, trained with a cosine learning rate decay~\cite{loshchilov2016sgdr} for 500 epochs with a maximum learning rate of $.40$, three epochs of linear learning rate warmup, a batch size of 512, momentum of $0.9$, and weight decay of $5 \cdot 10^{-4}$, obtaining a baseline accuracy of $90.0\%$.
Data augmentation consists of random crops and horizontal flips.
All experiments were done on two NVIDIA Tesla V100 GPUs with \verb|pytorch|~\cite{paszke2019pytorch}.

\begin{table*}
  \caption{The effect of each source of nondeterminism and several combinations of nondeterminism sources for ResNet-14 on CIFAR-10.
  The second and third columns give the standard deviation of accuracy and cross-entropy across 100 runs, varying only the nondeterminism source (700 trained models total).  Also given are error bars, corresponding to the standard deviation of each standard deviation.
  The fourth, fifth, and sixth columns give the average percentage of examples models disagree on, the average pairwise Spearman's correlation coefficient between predictions, and the average change in accuracy from ensembling two models, respectively (Sec.~\ref{sec:protocol}).
  }
  \label{tab:cifar10_nondeterminism}
  \centering
  \resizebox{.87\textwidth}{!}{\begin{tabular}{lccccc}
    \toprule
    \multirow{2}{*}{Nondeterminism Source}     & Accuracy &  Cross-Entropy & Pairwise & Pairwise & Ensemble\\
    & SD (\%) & SD  & Disagree (\%) & Corr. &  $\Delta$ (\%)\\
    \midrule
    Parameter Initialization      & $0.23 \pm 0.02$  & $0.0074 \pm 0.0005$ & 10.7 & 0.872 & 1.82   \\
    Data Shuffling                & $0.25 \pm 0.02$  & $0.0082 \pm 0.0005$ & 10.6 & 0.871 & 1.81   \\
    Data Augmentation             & $0.23 \pm 0.02$  & $0.0072 \pm 0.0005$ & 10.7 & 0.872 & 1.83   \\
    cuDNN                         & $0.22 \pm 0.01$  & $0.0083 \pm 0.0007$ & 10.5 & 0.873 & 1.76   \\
    Data Shuffling + cuDNN        & $0.21 \pm 0.01$  & $0.0077 \pm 0.0005$ & 10.6 & 0.871 & 1.80   \\
    Data Shuffling + Aug. + cuDNN & $0.22 \pm 0.01$  & $0.0074 \pm 0.0005$ & 10.7 & 0.871 & 1.84   \\
    All Nondeterminism Sources    & $0.26 \pm 0.02$  & $0.0072 \pm 0.0005$ & 10.7 & 0.871 & 1.82   \\
    \bottomrule
  \end{tabular}}
  \vspace{-5mm}
\end{table*}

We show the results of our protocol in this setting in Table~\ref{tab:cifar10_nondeterminism}.
Across all measures of performance variability and representation diversity, what we find is surprising and clear --- while there are slight differences, each source of nondeterminism has very similar effects on the variability of final trained models.
In fact, random parameter initialization, arguably the form of nondeterminism that variability in performance is most commonly attributed to, does not stand out based on any metric, and even combinations of multiple sources of nondeterminism produce remarkably little difference --- all are within a maximum of $20\%$ (relative) of each other.

\begin{figure}[t]
  \includegraphics[width=.78\linewidth]{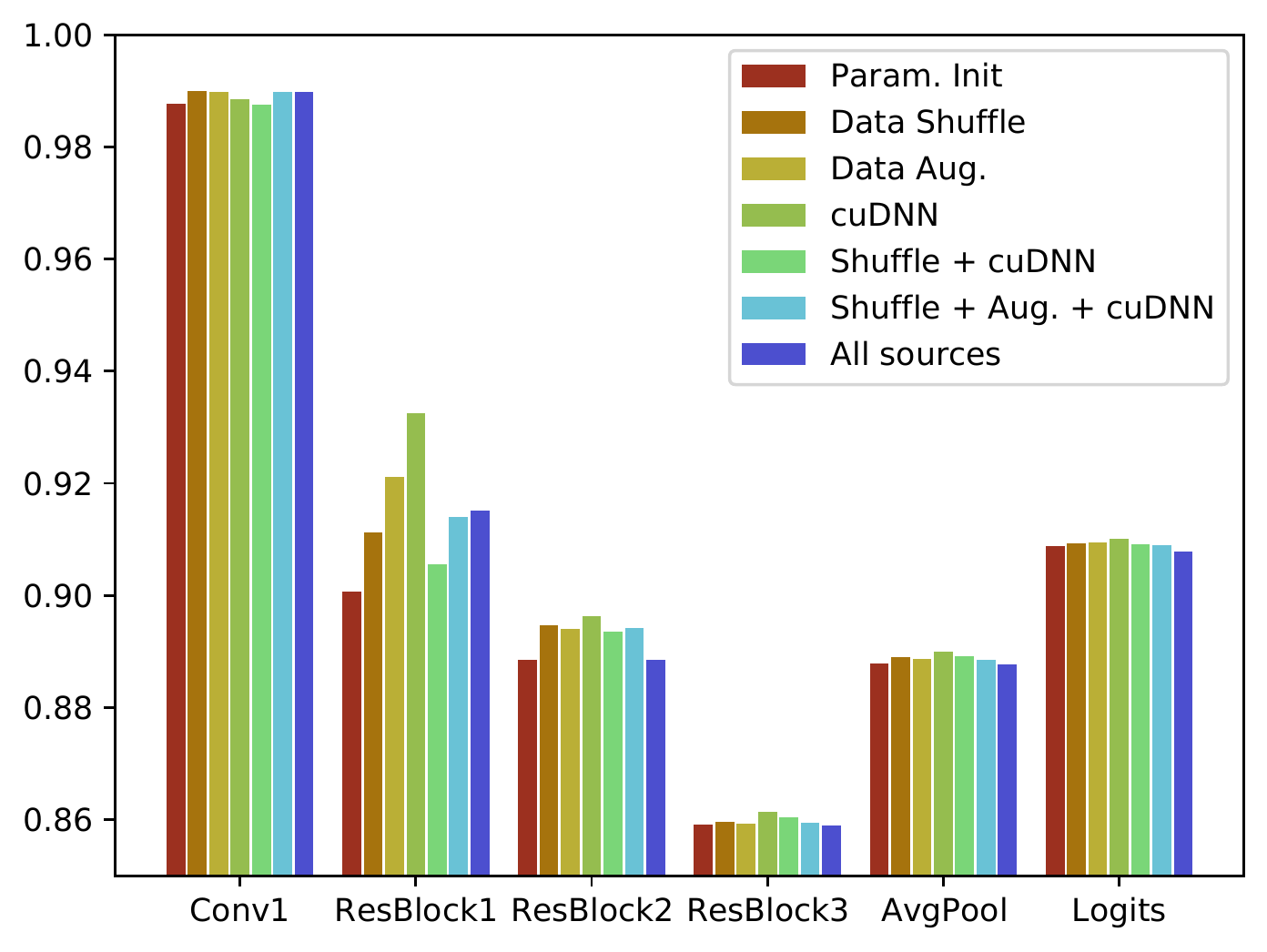}
  \caption{Average CKA representation similarity~\cite{kornblith2019similarity} for pairs of ResNet-14 models on CIFAR-10 across nondeterminism sources and a variety of network layers.
  }
  \label{fig:resnet14_ckas}
  \vspace{-5mm}
\end{figure}

Turning toward CKA and representational diversity on a per-layer level, we plot average CKA values across 6 representative layers in Fig.~\ref{fig:resnet14_ckas}, done for pairwise combinations of 25 models (due to the cost of CKA).
Consistent with other analyses, CKA reveals that while some differences in representational similarity exist between nondeterminism sources, particularly in the output of the first residual block, by and large these differences are small, easily dwarfed in size by representational differences across layers.

\begin{table*}
  \caption{The effect of each source of nondeterminism for a QRNN on Penn Treebank; 100 runs per row. Note that lower PPL is better for language modeling tasks, so changes in PPL from ensembling are negative.
  }
  \label{tab:ptb_nondeterminism}
  \centering
  \resizebox{.72\textwidth}{!}{\begin{tabular}{lccc}
    \toprule
    Nondeterminism Source         & PPL SD          & Pairwise Disagree (\%) & Ensemble PPL $\Delta$ \\
    \midrule
    Parameter Initialization      & $0.20 \pm 0.01$ &                 17.3  & -2.07\\
    Stochastic Operations         & $0.19 \pm 0.01$ &                 17.3  & -2.08 \\
    All Nondeterminism Sources    & $0.18 \pm 0.01$ &                 17.4  & -2.07 \\
    \bottomrule
  \end{tabular}}
  \vspace{-5mm}
\end{table*}

\subsection{Experiments in Language Modeling}
\label{sec:language_modeling}
Here we show that this phenomenon is not unique to image classification by applying the same experimental protocol to language modeling.
For these experiments, we employ a small quasi-recurrent neural network (QRNN)~\cite{bradbury2016quasi} on Penn Treebank~\cite{marcus1993building}, using the publicly available code of \cite{merity2017regularizing}.
This model uses a 256-dimensional word embedding, 512 hidden units per layer, and 3 layers of recurrent units, obtaining a perplexity (PPL) of 75.49 on the Penn Treebank test set.

For this task, two sources of nondeterminism are relevant: random parameter initialization, and stochastic operations, including a variation of dropout and variable length backpropagation through time, which share a common seed.
To measure performance variability, PPL is the most widely-accepted metric, and for diversity in representation we focus on only two metrics (pairwise disagreement and benefits from ensembling) because CKA was not designed for variable-length input and standard computing libraries~\cite{virtanen2020scipy} are not efficient enough to calculate $O(R^2)$ correlation coefficients with such large inputs.

We show results in Table~\ref{tab:ptb_nondeterminism}, where we find almost no difference across all diversity metrics, showing the phenomenon generalizes beyond image classification and ResNets.

\subsection{Nondeterminism Throughout Training}
One hypothesis for the this phenomenon's cause is the sensitivity of optimization in the initial phase of learning, which recent work has demonstrated in other contexts~\cite{achille2019critical,frankle2020early}.
With our experimental protocol, this is straightforward to test:
If this were the case, then training models identically for the first $N$ epochs and only then introducing nondeterminism would result in significantly less variability in final trained models, measured across all metrics.
Furthermore, by varying $N$, we can actually determine \emph{when} in training each source of nondeterminism has its effect (for sources that vary over the course of training, \emph{i.e.} not random parameter initialization).

\begin{figure}[t]
  \includegraphics[width=.90\linewidth]{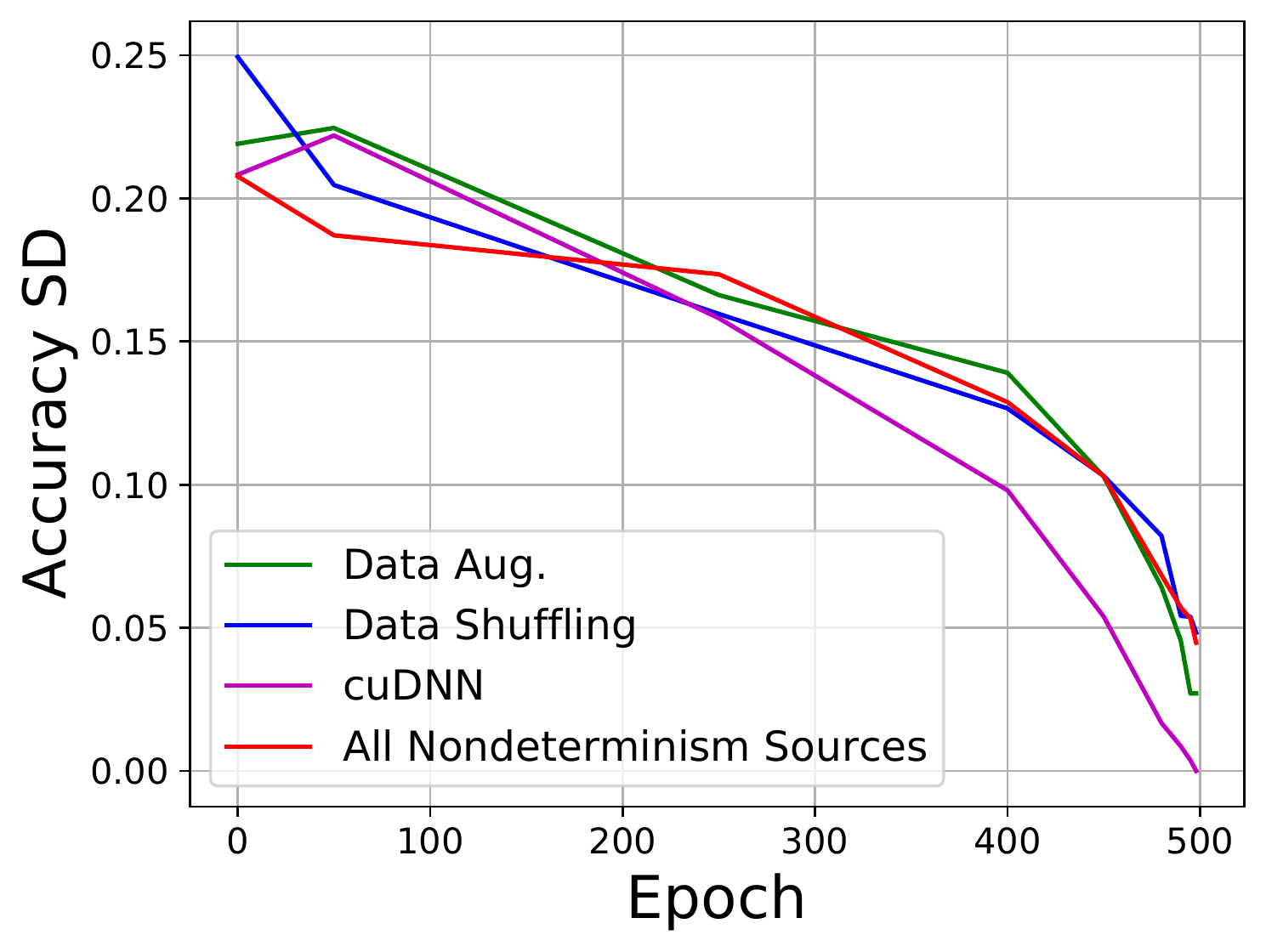}
  \caption{
    The effect of the onset of nondeterminism on the variability of accuracy in converged models.
    Each point corresponds to training 100 models deterministically for a certain number of epochs (x-axis), then enabling a given source of nondeterminism by varying its seed starting from that epoch and continuing through to the end of training, then measuring the accuracy SD (y-axis).
}
  \vspace{-5mm}
  \label{fig:over_time}
\end{figure}

We perform this experiment for the ResNet-14 model on CIFAR-10 in Fig.~\ref{fig:over_time}, where we find that the beginning of training is not particularly sensitive to nondeterminism.
Instead,  model variability is nearly as high when enabling nondeterminism even after 50 epochs, and we see only a gradual reduction in final model variability as the onset of nondeterminism is moved later and later.

\section{Instability}
\label{sec:instability}
Why does each source of nondeterminism have similar effects on model variability?
We approach this question by finding the smallest possible change that produces the same amount of variability.
In doing so, we find that only an extremely tiny change is necessary, thereby demonstrating the \emph{instability} in optimizing neural networks.

\subsection{Instability and Nondeterminism}
\label{sec:instability_nondeterminism}
To demonstrate, we perform a simple experiment:
First we deterministically train a simple ResNet-14 model on CIFAR-10, achieving a test cross-entropy of $0.3519$ and accuracy of $90.0\%$.
Then, we train another model in an identical fashion, with exactly equal settings for all sources of nondeterminism, but one extremely small change: we randomly pick a single weight in the first layer and change its value by the smallest possible amount in a 32-bit floating point representation, \emph{i.e.} an addition or subtraction of a single bit in the least-significant digit.
As an example, this could change a value from $-0.0066514308$ to $-0.0066514313$, a difference on the order of $5 \cdot 10^{-10}$.

What happens when we optimize this model, different from the original by only a single bit?
By the end of the first epoch of training, with the learning rate still warming up, the new model already differs in accuracy by $0.18\%$ ($25.74\%$ vs $25.56\%$).
In one more epoch the difference is a larger $2.33\%$ ($33.45\%$ vs $31.12\%$), and after three epochs, the difference is a staggering $10.42\%$ ($41.27\%$ vs $30.85\%$).
Finally, at the end of training the model weights converge, with the new model obtaining an accuracy of $90.12\%$ and a cross-entropy of $0.34335$, substantially different from the original despite only a tiny change in initialization.
Viewing the optimization process end-to-end, with the initial parameters as the input and a given performance metric as the output, this demonstrates a condition number $\frac{\|\delta f\|}{\|\delta x\|}$ of $1.8 \cdot 10^7$ for cross-entropy and $2.6 \cdot 10^8$ for accuracy.

\begin{table*}
  \caption{The effect of instability --- randomly changing a single weight by one bit during initialization for ResNet-14 on CIFAR-10.
  }
  \label{tab:cifar10_singleadd}
  \centering
  \begin{tabular}{lccccc}
    \toprule
    \multirow{2}{*}{Nondeterminism Source}     & Accuracy &  Cross-Entropy & Pairwise & Pairwise & Ensemble\\
    & SD (\%) & SD  & Disagree (\%) & Corr. &  $\Delta$ (\%)\\
    \midrule
    Random Bit Change & $0.21 \pm 0.01$ & $0.0068 \pm 0.0004$ & 10.6 & 0.874 & 1.82   \\
    \bottomrule
  \end{tabular}
  \vspace{-5mm}
\end{table*}

\begin{table*}
  \caption{The effect of instability for a QRNN on Penn Treebank. Also see Table~\ref{tab:ptb_nondeterminism} for comparison.
  }
  \label{tab:ptb_singleadd}
  \centering
  \begin{tabular}{lccc}
    \toprule
    Nondeterminism Source         & PPL SD          & Pairwise Disagree (\%) & Ensemble PPL $\Delta$ \\
    \midrule
    Random Bit Change & $0.19 \pm 0.01$ & 17.7 & -2.07  \\
    \bottomrule
  \end{tabular}
  \vspace{-5mm}
\end{table*}

We can more rigorously test this using our protocol from Sec.~\ref{sec:nondeterminism} --- this time, our source of nondeterminism is randomly picking a different weight to change in each model training run, then either incrementing or decrementing it to the next available floating-point value.
We show the results in Table~\ref{tab:cifar10_singleadd} for image classification on CIFAR-10 (\emph{c.f.} Table~\ref{tab:cifar10_nondeterminism} for comparison) and Table~\ref{tab:ptb_singleadd} for language modeling on Penn Treebank (\emph{c.f.} Table~\ref{tab:ptb_nondeterminism}), where we find that even this small change produces roughly as much variability in model performance as every other source of nondeterminism.

From this, it is easy to see why every other source of nondeterminism has similar effects --- so long as nondeterminism produces any change in model weights, whether by changing the input slightly, altering the gradient in some way, or any other effect, it will produce \emph{at least} as much model variability as caused by the instability of model optimization.

\subsection{Instability and Depth}
\label{sec:why_instable}
Instability occurs in networks of more than a single layer.

Due to convexity, linear models optimized with a cross-entropy loss and an appropriate learning rate schedule always converge to a global minimum.
However, in practice we find an even stronger property: when initial weights are modified by a single bit, beyond simply converging to the same final value, the entire optimization trajectory stays close to that of an unperturbed model, never differing by more than a vanishingly small amount.
At convergence, a set of linear models trained in this way with only single random bit changes had a final accuracy SD of 0 (\emph{i.e.} no changes in any test set predictions) and cross-entropy SD of ${\sim}1 \cdot 10^{-7}$, far below that of any deeper model.

In contrast, instability occurs as soon as a single hidden layer was added, with an accuracy SD of 0.28 and cross-entropy SD of 0.0051 for a model whose hidden layer is fully-connected, and an accuracy SD of 0.14 and cross-entropy SD of 0.0022 when the hidden layer is convolutional, both a factor of $10{,}000$ greater than the linear model.
See Appendix Secs.~\ref{sec:linear_twolayer} and \ref{sec:impact_of_random_bit_change_over_time} for full details and a visualization of the effects of instability during training.

\section{Reducing Variability}
\label{sec:reducing_variability}

Here we identify and demonstrate two approaches that partially mitigate the variability caused by nondeterminism and instability.
See the Appendix Sec.~\ref{sec:didnt_work} for learnings on approaches which were unsuccessful in reducing variability.

\subparagraph{Accelerated Ensembling.}
As previously mentioned, the standard practice for mitigating run-to-run variability is to train multiple independent copies of a model, gaining a more robust performance estimate by measuring a metric of interest over multiple trials.
Ensembling is a similar alternative approach, which shares the intuition of multiple independent training runs, but differs in that the predictions themselves are averaged and the performance of the ensembled model itself is measured.
Indeed, as demonstrated in Table~\ref{tab:cifar10_snapshot} (top), ensembles of larger models have less variability.
However, since ensembling still requires training multiple copies of models, is does not reduce the computational burden caused by nondeterminism and instability.

To that end, we propose the use of recent \emph{accelerated} ensembling techniques to reduce variability.
Accelerated ensembling is a new research direction in which only one training run is needed~\cite{huang2017snapshot,garipov2018loss,wen2020batchensemble}.
While such techniques typically underperform ensembles composed out of truly independent models, the nature of their accelerated training can reduce variability without incurring additional cost during training.
The approach we focus on is the Snapshot Ensemble~\cite{huang2017snapshot}, which uses a cyclic learning rate schedule, creating the members of an ensemble out of models where the learning rate is $0$ in the cyclic learning rate schedule.

In Table~\ref{tab:cifar10_snapshot} (bottom), we compare a snapshot ensemble (``Acc. Ens.'') with 5 cycles in its learning rate (\emph{i.e.} model snapshots are taken after every 100 epochs of training) to ordinary ensembling on CIFAR-10 with all sources of nondeterminism enabled.
Despite training only a single model, the accelerated ensemble had variability in accuracy and cross-entropy comparable to an ensemble of two independently-trained models, with other metrics comparable to those of even larger ensembles.
Across measures, accelerated ensembling reduces variability by an average of 48\% relative.

\begin{table*}
  \caption{
Comparison of single and ensemble model variability on CIFAR-10 with proposed methods for reducing the effects of nondeterminism.
For standard ensembles, $N$ denotes the number of constituent models, ``Acc. Ens.'' uses accelerated ensembling, and
[Single$\vert$Acc. Ens.]/[Flip$\vert$CropX$\vert$Flip-CropX]-TTA use either horizontal flips, crops (with $X$ crops), or flips and crops for test-time augmentation on top of either regular single models or an accelerated ensemble.
Also shown is the training time and average relative reduction in variability across metrics compared to the baseline `Single Model''.
All results are based on 100 runs of model training.
  }
  \label{tab:cifar10_snapshot}
  \centering
  \resizebox{\textwidth}{!}{\begin{tabular}{lccccccc}
    \toprule
    \multirow{2}{*}{Model}  & Training   & Accuracy &  Cross-Entropy & Pairwise & Pairwise & Ensemble & Variability \\
    & Cost & SD (\%) & SD  & Disagree (\%) & Corr. &  $\Delta$ (\%) & Reduction \\
    \midrule
    Single Model              & $1\times$  & $0.26 \pm 0.02$    & $0.0072 \pm 0.0005$     & 10.7 & 0.871 & 1.82 &  \emph{n/a} \\ 
    Ensemble ($N=2$)          & $2\times$  & $0.19 \pm 0.02$    & $0.0044 \pm 0.0004$     & 6.9  & 0.929 & 0.89 &  39\% \\
    Ensemble ($N=3$)          & $3\times$  & $0.15 \pm 0.02$    & $0.0033 \pm 0.0005$     & 5.5  & 0.951 & 0.59 &  55\% \\
    Ensemble ($N=4$)          & $4\times$  & $0.17 \pm 0.02$    & $0.0030 \pm 0.0004$     & 4.6  & 0.963 & 0.43 &  60\% \\
    Ensemble ($N=5$)          & $5\times$  & $0.12 \pm 0.02$    & $0.0028 \pm 0.0004$     & 4.1  & 0.970 & 0.34 &  67\% \\
    Ensemble ($N=10$)         & $10\times$ & $0.11 \pm 0.02$    & $0.0022 \pm 0.0004$     & 2.9  & 0.985 & 0.20 &  76\% \\
    Ensemble ($N=20$)         & $20\times$ & $0.11 \pm 0.04$    & $0.0018 \pm 0.0005$     & 2.0  & 0.992 & 0.08 &  81\% \\ \hline
    Acc. Ens.                 & $1\times$  & $0.19 \pm 0.02$    & $0.0044 \pm 0.0003$     & 6.1  & 0.957 & 0.63 &  48\% \\
    Single/Flip-TTA           & $1\times$  & $0.24 \pm 0.02$    & $0.0061 \pm 0.0005$     & 8.2  & 0.905 & 1.20 &  21\% \\
    Single/Crop25-TTA         & $1\times$  & $0.23 \pm 0.02$    & $0.0059 \pm 0.0004$     & 9.2  & 0.893 & 1.49 &  16\% \\
    Single/Crop81-TTA         & $1\times$  & $0.21 \pm 0.01$    & $0.0055 \pm 0.0004$     & 8.8  & 0.898 & 1.39 &  21\% \\
    Single/Flip-Crop25-TTA    & $1\times$  & $0.21 \pm 0.02$    & $0.0051 \pm 0.0004$     & 7.2  & 0.920 & 0.99 &  33\% \\
    Single/Flip-Crop81-TTA    & $1\times$  & $0.19 \pm 0.01$    & $0.0049 \pm 0.0004$     & 6.9  & 0.922 & 0.92 &  37\% \\
    Acc. Ens./Flip-TTA        & $1\times$  & $0.15 \pm 0.01$    & $0.0039 \pm 0.0003$     & 5.0  & 0.967 & 0.45 &  58\% \\
    Acc. Ens./Flip-Crop81-TTA & $1\times$  & $0.16 \pm 0.01$    & $0.0033 \pm 0.0002$     & 4.6  & 0.972 & 0.38 &  61\% \\
    \bottomrule
  \end{tabular}}
  \vspace{-5mm}
\end{table*}

\subparagraph{Test-Time Data Augmentation.}
Test-time data augmentation (TTA) is the practice of augmenting test set examples using data augmentation, averaging model predictions made on each augmented example, and is typically used to improve generalization~\cite{szegedy2015going}.
Beyond improved generalization, though, TTA can be thought of as a form of ensembling in data-space (as opposed to the model-space averaging of standard ensembling), giving it potential for mitigating the variability due to nondeterminism.

In Table~\ref{tab:cifar10_snapshot} (bottom), we show results on CIFAR-10 with horizontal flip and image cropping TTA (details in Appendix Sec.~\ref{sec:tta_details}), and also combine TTA with accelerated ensembling.
Simple flip TTA reduces variability across all metrics (21\% average relative reduction), standalone cropping reduces variability by 16\% to 21\% depending on the number of crops, and employing both pushes this up to 37\%.
Combined with accelerated model ensembling, variability is reduced by up 61\% without any increase in training budget.

\section{Generalization Experiments}
\label{sec:generalization_experiments}
In this section we detail additional experiments showing the generalization of our results on nondeterminism, instability, and methods for reducing variability to other datasets (MNIST, ImageNet) and model architectures.
We compile our main generalization results in Table~\ref{tab:generalization}, with additional results in the Appendix.

\begin{table*}
  \caption{Generalization experiments of nondeterminism and instability with other architectures on CIFAR-10, ImageNet, and MNIST. For CIFAR-10 and MNIST, each row is computed from the statistics of 100 trained models, and for ImageNet, each row is computed from 20 trained models.
  Within each section the most relevant comparisons to make are between ``Random Bit Change'' and ``All Nondeterminism Sources'' to evaluate instability, and between ``All Nondeterminism Sources'', ``Acc. Ens.'', and each TTA method to evaluate the efficacy of our proposals to mitigate the effects of nondeterminism and instability (all TTA models have all sources of nondeterminism enabled).
  Notation follows Tables~\ref{tab:cifar10_nondeterminism} and \ref{tab:cifar10_snapshot}, and all TTA cropping for CIFAR-10 uses the 81-crop variant.
  }
  \label{tab:generalization}
  \centering
  \resizebox{0.75\textwidth}{!}{\begin{tabular}{lrcccc}
    \toprule
    \multirow{2}{*}{Nondeterminism Source}     & Accuracy &  Cross-Entropy & Pairwise & Pairwise & Ensemble\\
    & SD (\%) & SD  & Disagree (\%) & Corr. &  $\Delta$ (\%)\\
    \midrule
    \multicolumn{2}{l}{CIFAR-10: ResNet-6} \\
    \cmidrule(r){1-1}
    Parameter Initialization   & $0.50 \pm 0.04$                & $0.0117 \pm 0.0010$                  & 20.0  & 0.925   & 2.17   \\
    All Nondeterminism Sources & $0.43 \pm 0.03$                & $0.0106 \pm 0.0007$                  & 20.1  & 0.924   & 2.17   \\
    Random Bit Change          & $0.41 \pm 0.02$                & $0.0094 \pm 0.0006$                  & 19.8  & 0.925   & 2.12   \\
    Single/Flip-Crop-TTA       & $0.44 \pm 0.03$                & $0.0096 \pm 0.0006$                  & 15.6  & 0.949   & 1.41   \\
    Acc. Ens.                  & $0.45 \pm 0.03$                & $0.0104 \pm 0.0007$                  & 14.0  & 0.963   & 0.99   \\
    Acc. Ens./Flip-Crop-TTA    & $0.43 \pm 0.03$                & $0.0096 \pm 0.0006$                  & 11.6  & 0.973   & 0.71   \\
    \midrule
    \multicolumn{2}{l}{CIFAR-10: ResNet-18} \\
    \cmidrule(r){1-1}
    Parameter Initialization   & $0.15 \pm 0.01$                & $0.0067 \pm 0.0005$                  &  4.7  & 0.814   & 0.71   \\
    All Nondeterminism Sources & $0.18 \pm 0.01$                & $0.0073 \pm 0.0005$                  &  4.8  & 0.808   & 0.75   \\
    Random Bit Change          & $0.13 \pm 0.01$                & $0.0060 \pm 0.0005$                  &  4.7  & 0.830   & 0.73   \\
    Single/Flip-Crop-TTA       & $0.14 \pm 0.01$                & $0.0047 \pm 0.0003$                  &  3.4  & 0.851   & 0.41   \\
    Acc. Ens.                  & $0.13 \pm 0.01$                & $0.0038 \pm 0.0003$                  &  2.9  & 0.884   & 0.31   \\
    Acc. Ens./Flip-Crop-TTA    & $0.11 \pm 0.01$                & $0.0029 \pm 0.0002$                  &  2.2  & 0.909   & 0.19   \\
    \midrule
    \multicolumn{2}{l}{CIFAR-10: ShuffleNetv2-50\%} \\
    \cmidrule(r){1-1}
    Parameter Initialization   & $0.22 \pm 0.01$                & $0.0112 \pm 0.0007$                  &  8.4  & 0.696   & 1.38   \\
    All Nondeterminism Sources & $0.22 \pm 0.02$                & $0.0123 \pm 0.0008$                  &  8.4  & 0.692   & 1.40   \\
    Random Bit Change          & $0.21 \pm 0.01$                & $0.0107 \pm 0.0006$                  &  8.3  & 0.695   & 1.36   \\
    Single/Flip-Crop-TTA       & $0.18 \pm 0.01$                & $0.0093 \pm 0.0007$                  &  6.5  & 0.762   & 0.90   \\
    Acc. Ens.                  & $0.18 \pm 0.01$                & $0.0067 \pm 0.0005$                  &  5.0  & 0.930   & 0.52   \\
    Acc. Ens./Flip-Crop-TTA    & $0.15 \pm 0.01$                & $0.0051 \pm 0.0004$                  &  4.1  & 0.948   & 0.35   \\
    \midrule
    \multicolumn{2}{l}{CIFAR-10: VGG-11} \\
    \cmidrule(r){1-1}
    Parameter Initialization   & $0.20 \pm 0.01$                & $0.0063 \pm 0.0004$                  &  6.6  & 0.807   & 0.91   \\
    All Nondeterminism Sources & $0.18 \pm 0.01$                & $0.0065 \pm 0.0004$                  &  6.6  & 0.806   & 0.94   \\
    Random Bit Change          & $0.16 \pm 0.01$                & $0.0060 \pm 0.0004$                  &  6.5  & 0.811   & 0.89   \\
    Single/Flip-Crop-TTA       & $0.15 \pm 0.01$                & $0.0042 \pm 0.0003$                  &  4.2  & 0.892   & 0.36   \\
    Acc. Ens.                  & $0.13 \pm 0.01$                & $0.0041 \pm 0.0003$                  &  4.1  & 0.914   & 0.39   \\
    Acc. Ens./Flip-Crop-TTA    & $0.11 \pm 0.01$                & $0.0026 \pm 0.0002$                  &  2.8  & 0.951   & 0.17   \\
    \midrule
    \multicolumn{2}{l}{MNIST} \\
    \cmidrule(r){1-1}
    Parameter Initialization   & $0.047 \pm 0.0036$             & $0.0024 \pm 0.0001$                  &  0.54 & 0.941   & 0.064  \\
    All Nondeterminism Sources & $0.046 \pm 0.0032$             & $0.0022 \pm 0.0001$                  &  0.56 & 0.939   & 0.068  \\
    Random Bit Change          & $0.035 \pm 0.0026$             & $0.0011 \pm 0.0001$                  &  0.30 & 0.989   & 0.011  \\
    Single/Crop-TTA            & $0.039 \pm 0.0025$             & $0.0016 \pm 0.0001$                  &  0.38 & 0.953   & 0.037  \\
    Acc. Ens.                  & $0.050 \pm 0.0031$             & $0.0019 \pm 0.0001$                  &  0.55 & 0.943   & 0.064  \\
    Acc. Ens./Crop-TTA         & $0.046 \pm 0.0028$             & $0.0013 \pm 0.0001$                  &  0.40 & 0.956   & 0.039  \\
    \midrule
    \multicolumn{2}{l}{ImageNet: ResNet-18} \\
    \cmidrule(r){1-1}
    All Nondeterminism Sources & $0.10 \pm 0.01$                & $0.0027 \pm 0.0004$                  &  20.7 & 0.814   & 1.94   \\
    Random Bit Change          & $0.09 \pm 0.01$                & $0.0026 \pm 0.0004$                  &  20.6 & 0.815   & 1.91   \\
    Single/Flip-TTA            & $0.12 \pm 0.02$                & $0.0022 \pm 0.0004$                  &  18.8 & 0.827   & 1.60   \\
    Single/Crop-TTA            & $0.10 \pm 0.02$                & $0.0023 \pm 0.0003$                  &  19.8 & 0.815   & 1.72   \\
    Single/Flip-Crop-TTA       & $0.11 \pm 0.01$                & $0.0018 \pm 0.0002$                  &  18.2 & 0.825   & 1.45   \\
    Acc. Ens.                  & $0.11 \pm 0.01$                & $0.0021 \pm 0.0003$                  &  14.4 & 0.919   & 0.80   \\
    Acc. Ens./Flip-Crop-TTA    & $0.09 \pm 0.01$                & $0.0018 \pm 0.0003$                  &  13.1 & 0.921   & 0.65   \\
    \bottomrule
  \end{tabular}}
\end{table*}

\subparagraph{CIFAR-10.}
On CIFAR-10, in addition to the ResNet-14 employed throughout this work, we experiment with a smaller 6-layer variant, larger 18-layer variant, VGG-11~\cite{simonyan2014very}, and a 50\%-capacity ShuffleNetv2~\cite{ma2018shufflenet}, with even more architectures in the Appendix.
As shown in Table~\ref{tab:generalization}, the observations around instability and its relationship to nondeterminism generally hold for these architectures, with a close correspondence between the magnitude of effects for a random bit change and each of the five metrics considered.

Turning towards our proposals (Sec.~\ref{sec:reducing_variability}) for mitigating the effects of nondeterminism and instability on model variability, we find across all model architectures that both accelerated ensembling and test-time augmentation reduce variability across nearly all metrics, with perhaps larger relative reductions for larger models and the pairwise metrics.
Only for the intersection of the smallest model (ResNet-6) and metrics of performance variability (Accuracy SD and Cross-Entropy SD) was there no benefit.

\subparagraph{MNIST.}
Experiments on MNIST~\cite{lecun1998gradient}, allow us to test whether our observations hold for tasks with very high accuracy --- $99.14\%$ for our relatively simple baseline model, which has two convolution and fully-connected layers.
As before, we find similar effects of nondeterminism for parameter initialization and all nondeterminism sources, including a comparable effect (albeit smaller) from a single random bit change, highlighting that the instability of training extends even to datasets where the goal is simpler and model performance is higher.
Of note, though, is the relative smaller effect of a single bit change on pairwise metrics of diversity, further suggesting that the magnitude of instability might be at least partially related to the interplay of model architecture, capacity, and degree of overfitting.

In terms of the mitigations against variability, only test-time augmentation appeared to significantly help.
For MNIST, the only augmentation employed was cropping, with a small 1-pixel padding (models were trained with no data augmentation).
While the fact that accelerated ensembling did not result in improvements is not particularly important in practice (since MNIST models are fast to train), it is an interesting result, which we hypothesize is also related to the degree of overfitting (similar to ResNet-6 on CIFAR-10).

\subparagraph{ImageNet.}
We perform larger-scale tests on ImageNet using 20 runs of a ResNet-18~\cite{he2016deep}, trained for 120 epochs, obtaining an average top-1 accuracy of 71.9\% on the ImageNet validation set.
Again, we find evidence supporting instability, with ``Random Bit Change'' having levels of variability comparable to models trained with all nondeterminism sources.
For reducing variability, we find modest improvements for both flipping-based and crop-based TTA on all metrics other than Accuracy SD, noting the large error bars of Accuracy and Cross-Entropy SD relative to their point estimates.
Accelerated ensembling follows a similar but stronger trend, with particularly large reductions in variability for pairwise metrics.

\section{Conclusion}
\label{sec:conclusion}
In this work, we have shown two surprising facts: First, though conventional wisdom holds that run-to-run variability in model performance is primarily determined by random parameter initialization, many sources of nondeterminism actually result in similar levels of variability.
Second, a key driver of this phenomenon is the instability of model optimization, in which changes on the order of $10^{-10}$ in a single weight at initialization can have as much effect as reinitializing all weights to completely random values.
We have also identified two approaches for reducing the variability in model performance and representation without incurring any additional training cost: ensembling in model-space via accelerated model ensembling, and ensembling in data-space via the application of test-time data augmentation.

Many promising directions for future work exist.
One important line of inquiry is in developing stronger theoretic understanding of the instability in optimization, beyond the largely empirical evidence in our work.
Another natural direction is improving upon the algorithms for reducing the effects of instability on model variability --- although both accelerated ensembling and TTA help, they are far from solving the problem entirely and incur additional computation during test time.
Last, it would be interesting to examine our findings on even larger models (\emph{e.g.} transformers for NLP and image recognition) and problems outside the fully supervised setting.
We hope that our work has shed light on a complex phenomenon that affects all deep learning researchers and inspires further research.

\clearpage

\bibliography{references}
\bibliographystyle{icml2021}

\appendix
\clearpage

\section{Linear, 2-Layer, and ResNet-10 Results}
\label{sec:linear_twolayer}
In Table~\ref{tab:linear_onehidden} we include results for linear networks, 2-layer networks (1 hidden layer), and a ResNet-10 on CIFAR-10, omitted from the main text due to space constraints.
As noted in the main text, parameter initialization generally has less effect for linear models, and random bit changes in particular have nearly no effect on linear models, highlighting the stability of SGD in optimizing linear models.
Also of note is the relative smaller effect of a single bit change for a 2-layer network where the hidden layer is convolutional --- still much larger than for the linear model, but significantly smaller than for any other non-linear model.
This suggests a similar effect to what was previously observed on MNIST (Table~\ref{tab:generalization}) in that degree of instability might be related to the interplay of model, dataset, and the degree of overfitting.

Otherwise, these results follow the results in the main text, wherein each source of nondeterminism has roughly the same effect as each other, and a majority of this is due to the instability of optimization, evidenced by the high variability in models with only random bit changes at initialization.
Test-time augmentation also remains effective in reducing model variability as compared to ``All Nondeterminism Sources'', the setting TTA is applied to.

\begin{table*}
  \caption{Linear, 2-layer, and ResNet-10 experiments on CIFAR-10.}
  \label{tab:linear_onehidden}
  \centering
  \resizebox{0.80 \textwidth}{!}{\begin{tabular}{lccccc}
    \toprule
    \multirow{2}{*}{Nondeterminism Source}     & Accuracy &  Cross-Entropy & Pairwise & Pairwise & Ensemble\\
    & SD (\%) & SD  & Disagree (\%) & Corr. &  $\Delta$ (\%)\\
    \midrule
    \multicolumn{2}{l}{CIFAR-10: Linear model} \\
    \cmidrule(r){1-1}
    Parameter Initialization   & $0.03 \pm \text{2e-3}$         & $0.0002 \pm \text{1e-5}     $        &  0.5 & 0.997   & \text{-4e-3}   \\
    All Nondeterminism Sources & $0.10 \pm 0.01$                & $0.0007 \pm \text{4e-5}     $        &  5.5 & 0.996   & 0.06   \\
    Random Bit Change          & $0.00 \pm 0.00$                & $\text{1e-7} \pm \text{1e-8} $       &  0.0 & 1.000   & 0.00   \\
    Single/Flip-Crop-TTA       & $0.05 \pm \text{3e-3}$         & $0.0001 \pm \text{1e-5}$             &  0.9 & 0.998   & \text{-3e-3}   \\
    \midrule
    \multicolumn{2}{l}{CIFAR-10: One hidden layer (fully-connected)} \\
    \cmidrule(r){1-1}
    Parameter Initialization   & $0.31 \pm 0.02$                & $0.0051 \pm 0.0003$                  & 24.2 & 0.941   & 1.38   \\
    All Nondeterminism Sources & $0.30 \pm 0.02$                & $0.0054 \pm 0.0004$                  & 24.9 & 0.937   & 1.49   \\
    Random Bit Change          & $0.28 \pm 0.02$                & $0.0051 \pm 0.0004$                  & 23.4 & 0.945   & 1.32   \\
    Single/Flip-Crop-TTA       & $0.15 \pm 0.01$                & $0.0017 \pm 0.0001$                  &  7.1 & 0.993   & 0.15   \\
    \midrule
    \multicolumn{2}{l}{CIFAR-10: One hidden layer (convolutional)} \\
    \cmidrule(r){1-1}
    Parameter Initialization   & $0.26 \pm 0.02$                & $0.0040 \pm 0.0003$                  & 12.5 & 0.974   & 0.64   \\
    All Nondeterminism Sources & $0.22 \pm 0.01$                & $0.0042 \pm 0.0003$                  & 12.7 & 0.973   & 0.68   \\
    Random Bit Change          & $0.14 \pm 0.01$                & $0.0022 \pm 0.0002$                  &  6.4 & 0.993   & 0.18   \\
    Single/Flip-Crop-TTA       & $0.19 \pm 0.01$                & $0.0033 \pm 0.0002$                  &  7.5 & 0.989   & 0.24   \\
    \midrule
    \multicolumn{2}{l}{CIFAR-10: ResNet-10} \\
    \cmidrule(r){1-1}
    Parameter Initialization   & $0.23 \pm 0.01$                & $0.0060 \pm 0.0003$                  & 13.7  & 0.912   & 2.13   \\
    All Nondeterminism Sources & $0.23 \pm 0.01$                & $0.0065 \pm 0.0004$                  & 13.6  & 0.911   & 2.13   \\
    Random Bit Change          & $0.25 \pm 0.02$                & $0.0065 \pm 0.0005$                  & 13.5  & 0.913   & 2.08   \\
    Single/Flip-Crop-TTA       & $0.24 \pm 0.02$                & $0.0047 \pm 0.0003$                  &  9.5  & 0.943   & 1.22   \\
    Acc. Ens.                  & $0.23 \pm 0.01$                & $0.0047 \pm 0.0003$                  &  8.8  & 0.962   & 0.96   \\
    Acc. Ens./Flip-Crop-TTA    & $0.18 \pm 0.01$                & $0.0035 \pm 0.0002$                  &  6.7  & 0.973   & 0.58   \\
    \bottomrule
  \end{tabular}}
  \vspace{-5mm}
\end{table*}

\section{Impact of Random Bit Changes Over Time}
\label{sec:impact_of_random_bit_change_over_time}

\begin{figure*}
  \includegraphics[width=.45\linewidth]{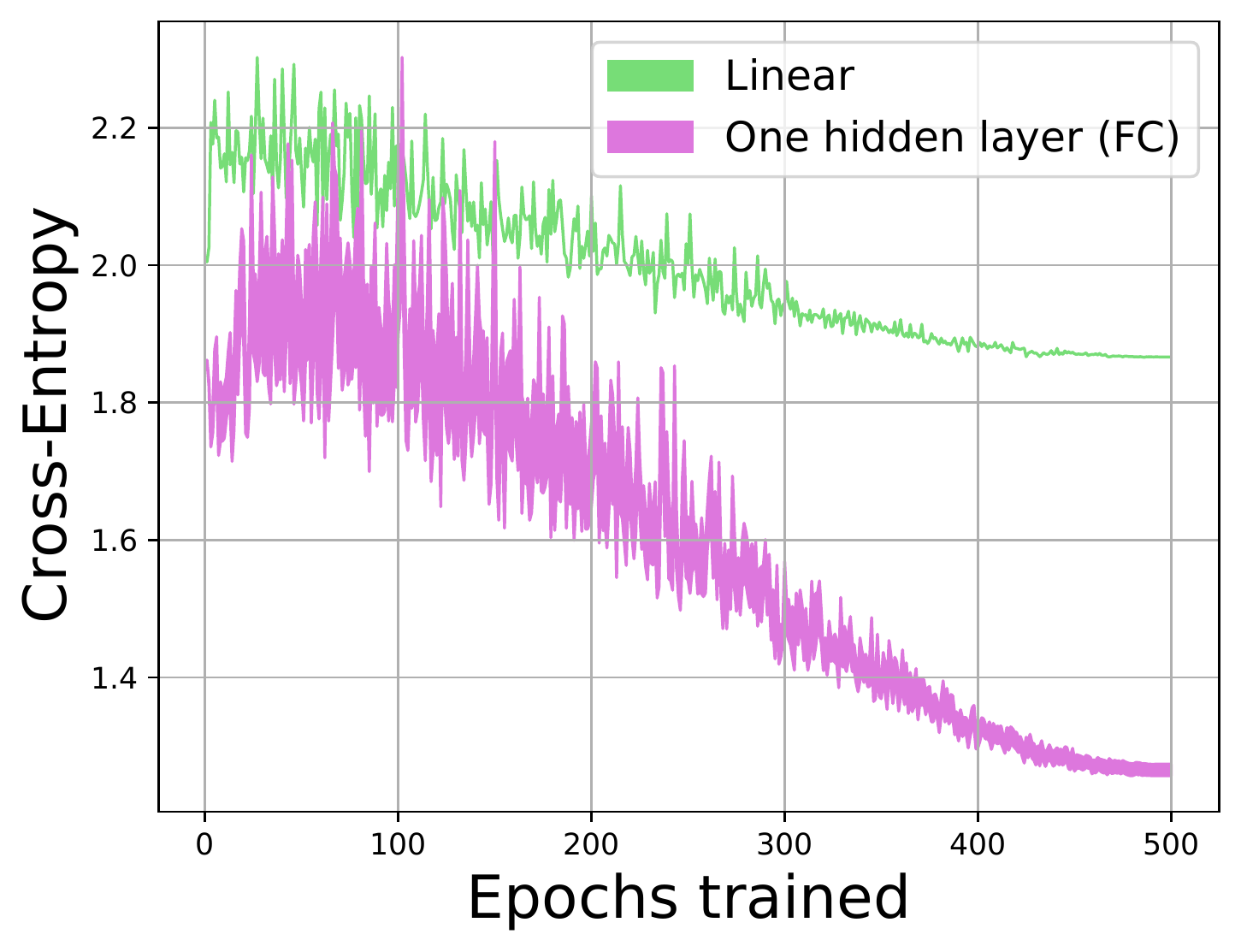}
  \quad
  \includegraphics[width=.45\linewidth]{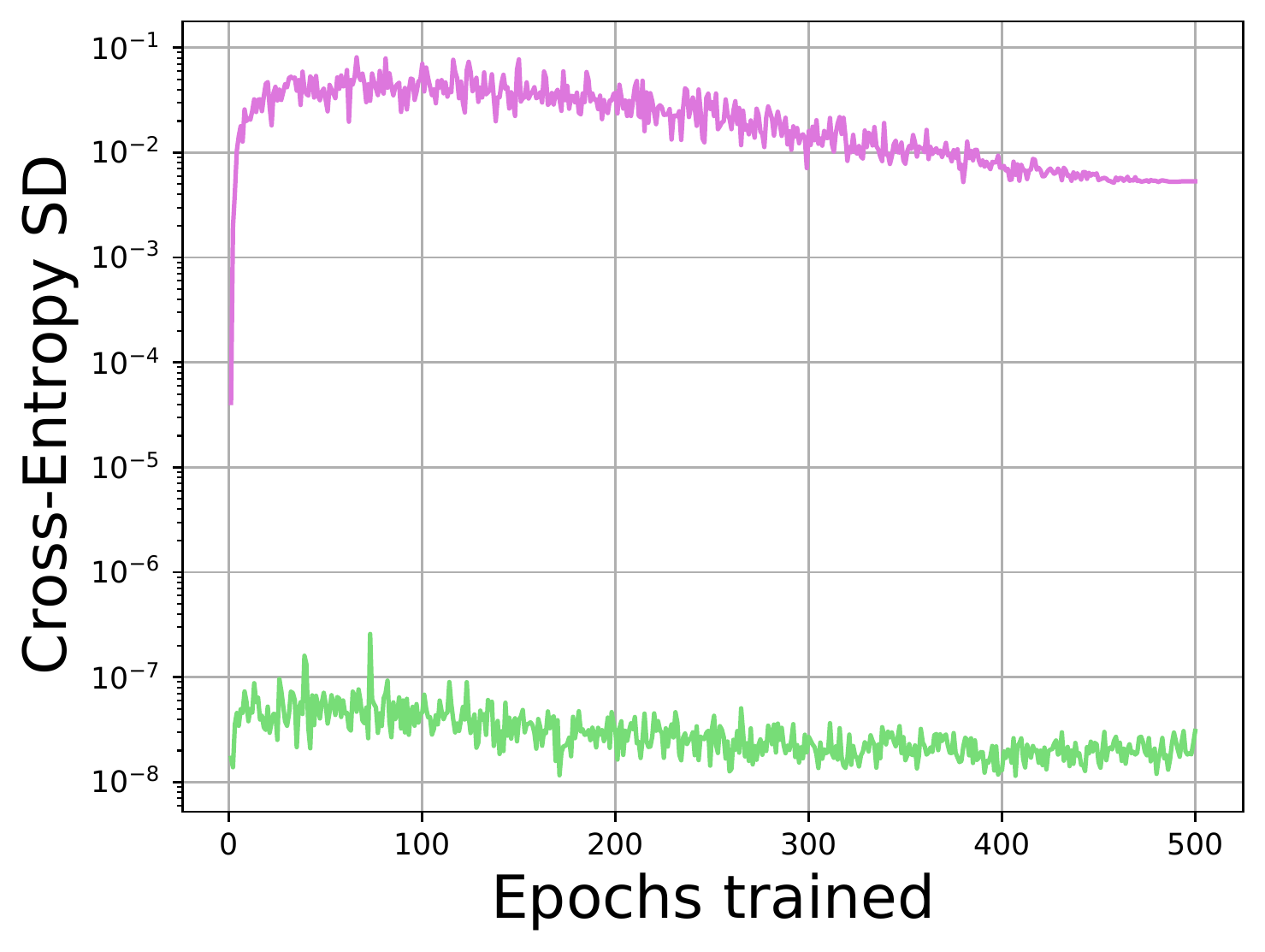}

  \includegraphics[width=.45\linewidth]{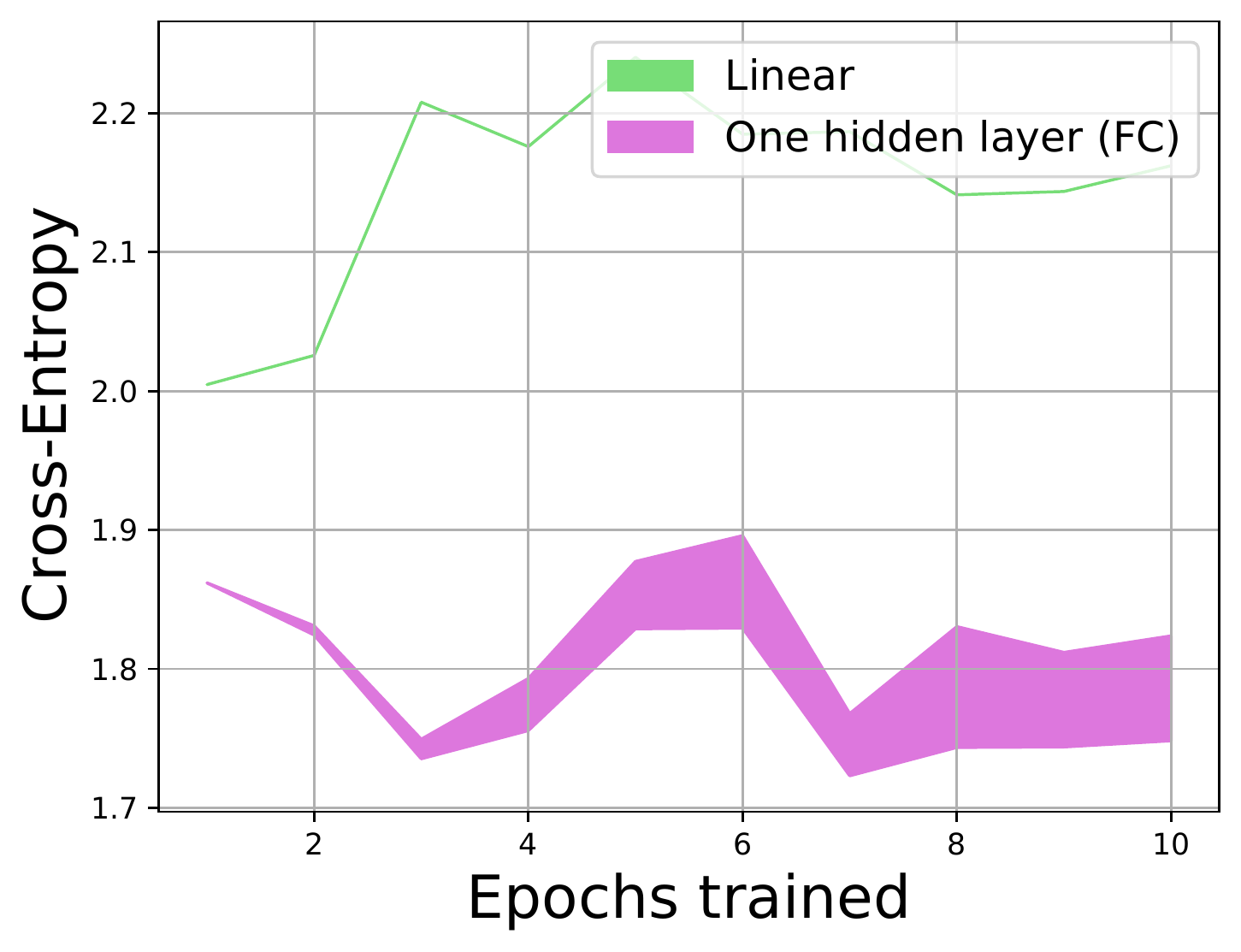}
  \quad
  \includegraphics[width=.45\linewidth]{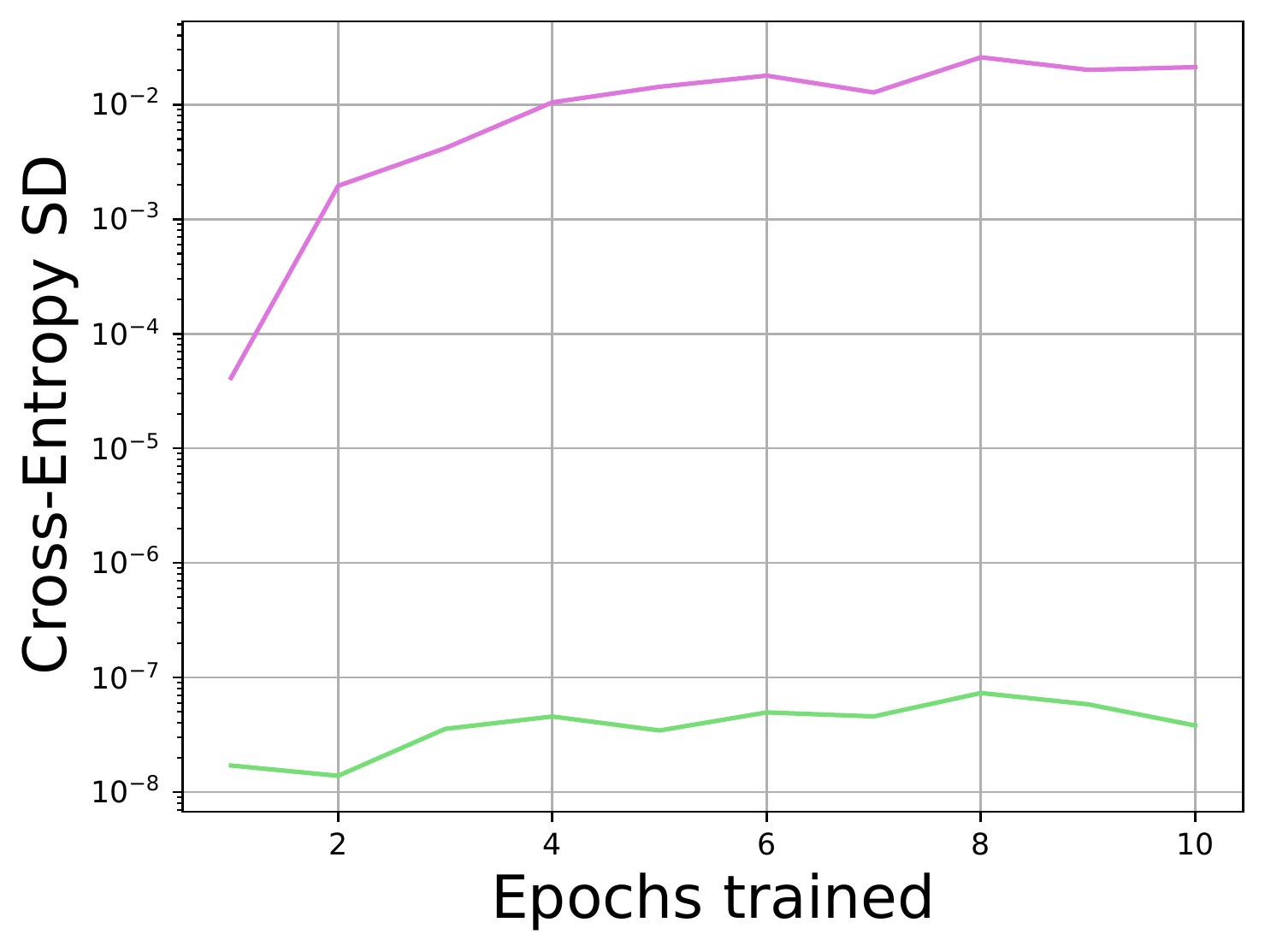}
  \caption{
    The impact of a random bit change during initialization for linear models vs 2-layer models with a single fully-connected hidden layer, where row 1 considers the full 500 epochs of training, and row 2 zooms in on the first 10 epochs.
    The left column of each row gives the range of cross-entropy values for 100 models in the middle 95th percentile of cross-entropy, plotted at each epoch.
    The right column of each row presents the standard deviation of these models, roughly corresponding to half the width of the corresponding range in the left-hand plots.
}
  \label{fig:singleadd_over_time}
  \vspace{-5mm}
\end{figure*}
In Fig.~\ref{fig:singleadd_over_time} we plot the effect of a random bit change for a linear and single hidden layer model on CIFAR-10, illustrating the effect of instability as described in Sec.~\ref{sec:instability}.
In the first few epochs of training, we observe that the standard deviation and range of cross-entropy for the model with one hidden layer quickly grows, only eventually decreasing much later in training as the model's parameters converges toward their final values.
On the other hand, for linear models, the standard deviation consistently remains 5 or more orders of magnitude lower throughout training.

\section{Impact of Network Width}
Here we present results studying the interaction of instability and model width, training variants of the ResNet-14 on CIFAR-10 with width modifiers ranging from $0.125$ to $8$ in powers of two.
Results are given in Table~\ref{tab:resnet14_width}.
Across model widths, we observe that instability still produces roughly as much variability as all other nondeterminism sources combined, as shown by the relative variability of models trained with ``Random Bit Change'' as a source of nondeterminism compared with those trained with all other sources (``All Sources'').
The only minor outlier was the width multiplier of $0.125$, for which random bit changes produced somewhat less variability, reminiscent of the experiments on MNIST (Table~\ref{tab:generalization}).
We also observe that variability generally decreases as model width increases (except for ``Pairwise Corr.''), though it is difficult to say whether this is due to a unique property of model width or simply wider models having higher test set performance.

\begin{table*}
  \caption{Experiments varying model width for ResNet-14 on CIFAR-10. In each row, the experimental setting is abbreviated by [width multiplier]/[sources of nondeterminism] (\emph{N=}[number of models trained]).
  }
  \label{tab:resnet14_width}
  \centering
  \small
  \resizebox{0.90 \textwidth}{!}{\begin{tabular}{lccccc}
    \toprule
    \multirow{2}{*}{Setting}     & Accuracy &  Cross-Entropy & Pairwise & Pairwise & Ensemble\\
    & SD (\%) & SD  & Disagree (\%) & Corr. &  $\Delta$ (\%)\\
    \midrule
    $0.125$/All Sources (\emph{N=30})                 & $0.86 \pm 0.10$ & $0.0183 \pm 0.0020$ & 40.8 & 0.887 & 2.04 \\
    $0.125$/Random Bit Change (\emph{N=30})           & $0.66 \pm 0.10$ & $0.0133 \pm 0.0020$ & 38.0 & 0.943 & 1.73 \\
    $0.25$/All Sources (\emph{N=30})                  & $0.59 \pm 0.06$ & $0.0124 \pm 0.0012$ & 25.9 & 0.880 & 2.37 \\
    $0.25$/Random Bit Change (\emph{N=30})            & $0.71 \pm 0.08$ & $0.0145 \pm 0.0019$ & 25.2 & 0.908 & 2.15 \\
    $0.5$/All Sources (\emph{N=30})                   & $0.33 \pm 0.04$ & $0.0077 \pm 0.0010$ & 15.8 & 0.890 & 2.21 \\
    $0.5$/Random Bit Change (\emph{N=30})             & $0.30 \pm 0.04$ & $0.0076 \pm 0.0008$ & 15.0 & 0.917 & 1.98 \\
    $1.0$/All Sources (\emph{N=100})                  & $0.26 \pm 0.02$ & $0.0072 \pm 0.0005$ & 10.7 & 0.871 & 1.82 \\
    $1.0$/Random Bit Change (\emph{N=100})            & $0.21 \pm 0.01$ & $0.0068 \pm 0.0004$ & 10.6 & 0.874 & 1.82 \\
    $2.0$/All Sources (\emph{N=30})                   & $0.13 \pm 0.01$ & $0.0066 \pm 0.0008$ &  7.1 & 0.790 & 1.19 \\
    $2.0$/Random Bit Change (\emph{N=30})             & $0.18 \pm 0.02$ & $0.0054 \pm 0.0012$ &  7.0 & 0.798 & 1.13 \\
    $4.0$/All Sources (\emph{N=30})                   & $0.14 \pm 0.02$ & $0.0049 \pm 0.0006$ &  5.0 & 0.781 & 0.75 \\
    $4.0$/Random Bit Change (\emph{N=30})             & $0.15 \pm 0.02$ & $0.0058 \pm 0.0007$ &  4.9 & 0.785 & 0.74 \\
    $8.0$/All Sources (\emph{N=30})                   & $0.13 \pm 0.02$ & $0.0038 \pm 0.0005$ &  3.8 & 0.807 & 0.53 \\
    $8.0$/Random Bit Change (\emph{N=30})             & $0.11 \pm 0.01$ & $0.0039 \pm 0.0004$ &  3.8 & 0.809 & 0.50 \\
    \bottomrule
  \end{tabular}}
  \vspace{-5mm}
\end{table*}

\section{Test-Time Augmentation Details}
\label{sec:tta_details}
\subparagraph{CIFAR-10.}
On CIFAR-10, in addition to TTA with horizontal flipping (\emph{i.e.} ensembling model predictions on the original image with its horizontally-flipped version), we also used a form of TTA with cropping.
Our usage of crop-based TTA was based on the version of cropping used as data augmentation during model training, in which each image was zero-padded by four pixels along each side, after which a random $32\times32$ crop was drawn.
In the main text, we experimented with 25- and 81-crop TTA variants, where the 81-crop variant uses all possible crops [$(4 \cdot 2 + 1)^2 = 81$], and the 25-crop variant uses a stride of 2 when sampling the possible crops [$(2 \cdot 2 + 1)^2 = 25$].
When adding in horizontally-flipped versions, the number of TTA examples doubles, for a maximum of 162 augmented versions of the original image.

\subparagraph{ImageNet.}
On ImageNet, the standard evaluation protocol we use for our experiments first resizes each image to have its smaller side be length 256, after which the central $224 \times 224$ crop is taken.
For TTA, besides the horizontal image flipping used in CIFAR-10, we also experimented with crops (`Crop-TTA' and `Flip-Crop-TTA') as follows: after each image is resized to have its smaller side length 256, a central $256 \times 256$ crop is taken, and then 9 crops of size $224 \times 224$ are taken in a $3 \times 3$ grid, starting from the top-left, and where the spacing between crops in the grid is 16 pixels.

\section{Approaches That Don't Reduce Instability}
\label{sec:didnt_work}
In the process of finding an approach that reduces run-to-run variability of models (Sec.~\ref{sec:reducing_variability}), we experimented with many approaches which all failed to make a dent in improving variability and stability.
For the benefit of the field, here we provide our experiences with these approaches which did not succeed in improving stability, despite the intuitive arguments for why they might help.

\subparagraph{Learning Rate and Duration of Training.} Noticing that the effects of nondeterminism seemed to accumulate during the course of training (Fig.~\ref{fig:over_time}), it seemed reasonable that varying the learning rate or duration of training might have an effect. However, varying the duration of training from anywhere between 50 and $2{,}000$ epochs on CIFAR-10 all produced models with a similar variance in performance as the results in the rest of this work (which used 500 epochs), even though the absolute performance differed by up to ${\sim}2\%$.

We show these results in Table~\ref{tab:lr_epoch}.
In general, increasing the number of epochs or changing the learning rate did not change the variability in performance (Accuracy SD; Cross-Entropy SD) much, with only a very slight increase in variability as the number of epochs grew to extremely large values (\emph{i.e.} 2,000 epochs).
There were slightly larger changes in pairwise representation-based metrics, where training longer again increased run-to-run variability.
However, none of these attempts actually reduced variability while maintaining performance; they only served to potentially make it larger.

As part of these experiments, we also verified the effects of instability with only 200 epochs of training and the effectiveness of accelerated ensembling techniques (``Acc. Ens.'') with this reduced training time, given in the last three rows of Table~\ref{tab:lr_epoch}.

\subparagraph{Choice of Optimizer.} Since instability and nondeterminism are both a property of optimization, it is conceivable that use of a different optimizer might be able to lessen the degree of instability in model training.
We experimented with SGD using various values of momentum, ranging from 0 for pure SGD to $0.999$ for a momentum-driven optimizer, but none succeeded in reduce instability. In addition, we experimented with Adam~\citep{kingma2014adam}, picked as a representative of the class of adaptive learning rate algorithms, but this, too, had no effect on stability.

\begin{table*}
  \caption{Experiments varying the learning rate and number of epochs for ResNet-14 on CIFAR-10. In each row, the experimental setting is abbreviated by [sources of nondeterminism]/[maximum learning rate]/[number of epochs] (\emph{N=}[number of models trained]), with the exception of the last row, which is a Snapshot ensemble but otherwise follows the same format.
  }
  \label{tab:lr_epoch}
  \centering
  \small
  \resizebox{\textwidth}{!}{\begin{tabular}{lccccc}
    \toprule
    \multirow{2}{*}{Setting}     & Accuracy &  Cross-Entropy & Pairwise & Pairwise & Ensemble\\
    & SD (\%) & SD  & Disagree (\%) & Corr. &  $\Delta$ (\%)\\
    \midrule
    All Sources/.40/50 (\emph{N=20})                  & $0.31 \pm 0.04$ & $0.0070 \pm 0.0007$ & 11.4 & 0.922 & 1.54 \\
    All Sources/.40/100 (\emph{N=20})                 & $0.26 \pm 0.04$ & $0.0068 \pm 0.0015$ & 10.8 & 0.909 & 1.71 \\
    All Sources/.40/250 (\emph{N=20})                 & $0.19 \pm 0.03$ & $0.0054 \pm 0.0009$ & 10.7 & 0.889 & 1.78 \\
    All Sources/.40/500 (\emph{N=100})                & $0.26 \pm 0.02$ & $0.0072 \pm 0.0005$ & 10.7 & 0.871 & 1.82 \\
    All Sources/.40/2000 (\emph{N=20})                & $0.24 \pm 0.02$ & $0.0096 \pm 0.0013$ & 11.2 & 0.828 & 2.08 \\
    Shuffle/.40/500 (\emph{N=100})                    & $0.25 \pm 0.02$ & $0.0082 \pm 0.0005$ & 10.6 & 0.871 & 1.81 \\
    Shuffle/.20/500 (\emph{N=100})                    & $0.23 \pm 0.02$ & $0.0071 \pm 0.0005$ & 11.0 & 0.858 & 1.95 \\
    Shuffle/.20/1000 (\emph{N=100})                   & $0.21 \pm 0.02$ & $0.0088 \pm 0.0005$ & 11.1 & 0.837 & 2.02 \\
    Shuffle/.10/500 (\emph{N=100})                    & $0.20 \pm 0.01$ & $0.0076 \pm 0.0005$ & 11.6 & 0.845 & 2.08 \\
    Shuffle/.10/2000 (\emph{N=100})                   & $0.24 \pm 0.02$ & $0.0100 \pm 0.0006$ & 11.6 & 0.801 & 2.19 \\
    Param. Init/.40/500 (\emph{N=100})                & $0.23 \pm 0.02$ & $0.0074 \pm 0.0005$ & 10.7 & 0.872 & 1.82 \\
    Param. Init/.20/500  (\emph{N=100})               & $0.23 \pm 0.02$ & $0.0084 \pm 0.0005$ & 11.0 & 0.859 & 1.97 \\
    Param. Init/.20/1000 (\emph{N=100})               & $0.25 \pm 0.02$ & $0.0095 \pm 0.0007$ & 11.1 & 0.836 & 2.06 \\
    Param. Init/.10/500 (\emph{N=100})                & $0.26 \pm 0.02$ & $0.0083 \pm 0.0005$ & 11.7 & 0.844 & 2.13 \\
    Param. Init/.10/2000 (\emph{N=100})               & $0.22 \pm 0.01$ & $0.0093 \pm 0.0008$ & 11.6 & 0.800 & 2.18 \\
    All Sources/.40/200 (\emph{N=100})                & $0.23 \pm 0.02$ & $0.0076 \pm 0.0004$ & 10.6 & 0.895 & 1.75 \\
    Random Bit/.40/200 (\emph{N=100})                 & $0.21 \pm 0.01$ & $0.0067 \pm 0.0004$ & 10.3 & 0.897 & 1.70 \\
    Acc. Ens./All Sources/.40/200 (\emph{N=100})      & $0.21 \pm 0.01$ & $0.0046 \pm 0.0003$ &  6.6 & 0.963 & 0.68 \\
    \bottomrule
  \end{tabular}}
  \vspace{-5mm}
\end{table*}

\subparagraph{Aggressive Stochastic Weight Averaging.} Inspired by the success found by \citet{madhyastha2019model}, we tried Aggressive Stochastic Weight Averaging (ASWA), a variant of SWA~\citep{izmailov2018averaging}. However, we could not get the model to converge to a reasonable degree of performance with the original formulation due to update sizes that decreased too rapidly, and though we were able to modify it to converge successfully, the output variance remained as high as the other models.

\subparagraph{Gradient Clipping.} With the intuition that instability might be caused by spurious gradients of large magnitude, we experimented with clipping the norm of gradients (using \verb|pytorch|'s implementation of \verb|torch.nn.utils.clip_grad.clip_grad_norm_|. Like other approaches, though, this had no effect on model variability.

\subparagraph{Weight Augmentation.} A very experimental approach, to reduce instability we experimented with taking an averaged gradient around the current set of parameters at each step, approximated by sampling a random weight offset before doing a forward or backward pass through the model. Intuitively, this might encourage optimization to not be too sensitive to the current value of weights; however, in practice this simply didn't affect the variance or stability of the model.

\section{Accelerated Ensembling in Language Modeling}
Although the accelerated ensembling technique we employed in the main text, Snapshot Ensembles~\cite{huang2017snapshot}, was only designed for image classification, we have also experimented with its usage for language modeling (see Sec.~\ref{sec:language_modeling} for problem setup).
We present results in Table~\ref{tab:ptb_nondeterminism_snap}, where we additionally compare models trained for 500 and 1000 epochs.
In both cases, accelerated ensembling resulted in lower model variability when considering pairwise metrics (reducing the fraction of tokens models disagreed on and reducing the PPL improvement from ensembling).
However, the variability in PPL was more mixed, and in fact we note that the accelerated ensembles actually had higher average PPL than their counterparts (\emph{e.g.} 75.0 for the accelerated ensemble vs 73.0 for the regular model), indicating that alternative accelerated ensembling techniques may be warranted for language modeling.

\begin{table*}
  \caption{The effects of accelerated model ensembling on Penn Treebank; 100 runs per row.
  ``Acc. Ens.'' indicates accelerated ensembling, and the trailing number in each setting name is the number of epochs models are trained for.
  }
  \label{tab:ptb_nondeterminism_snap}
  \centering
  \resizebox{.75\textwidth}{!}{\begin{tabular}{lccc}
    \toprule
    Setting         & PPL SD          & Pairwise Disagree (\%) & Ensemble PPL $\Delta$ \\
    \midrule
    All Nondeterminism Sources/500    & $0.18 \pm 0.01$ &                 17.4  & -2.07 \\
    Acc. Ens./All Sources/500         & $0.21 \pm 0.02$ &                 13.7  & -1.33 \\
    All Nondeterminism Sources/1000   & $0.17 \pm 0.01$ &                 17.6  & -2.08 \\
    Acc. Ens./All Sources/1000        & $0.16 \pm 0.01$ &                 14.1  & -1.34 \\
    \bottomrule
  \end{tabular}}
  \vspace{-5mm}
\end{table*}

\section{Subtleties in Evaluation}
\paragraph{Choice of Seeds.}
While we have done our best to make our experimental protocol straightforward and easy to interpret, one subtlety related to seed selection arises when examining results.
First, recall that in our example from Sec.~\ref{sec:protocol}, testing the effects of random initialization corresponded to training models for $(S_1, S_2, S_3) \in \{(i, 1, 1)\}_{i=1}^{R}$, where $S_1$ was the seed for random initialization, $S_2$ was the seed for training data shuffling, and $S_3$ was set to 1 to indicate the deterministic mode for cuDNN.
The subtlety arises in that the resulting distribution of $(S_1, S_2, S_3) \in \{(i, 1, 1)\}_{i=1}^{R}$ is not necessarily the same as the distribution where $S_2$ is set to a different arbitrary constant value, \emph{e.g.} $S_2 = 2$.
Due to this, there may be minor discrepancies when comparing the diversity in performance between two different sources of nondeterminism (though unlikely to change general conclusions unless the magnitude of the discrepancy is very large).
Combined with the natural sampling variability implicit in only training a finite number of models, this can lead to paradoxical results such as the standard deviation for a particular metric being slightly higher for a random bit change as compared to an entirely different random parameter initialization.
While we have separately validated that the general conclusions of our results hold when varying a few of these constant factors (\emph{i.e.} running experiments where $S_2$ is set to $2$ and $3$, in this example), it is difficult to resolve the discrepancy entirely without models according to the full cross-product of random seeds, which is prohibitive due to the exponential amount of required computation.

\paragraph{Variability vs Model Performance.}
Another challenge when interpreting results is that model variability covaries with model performance, which impacts validating any approach that affects both model variability and performance.
Though this is a difficult evaluation issue to solve in general, it is possible to show that the reductions in variability from Accelerated Ensembling and Test-Time Augmentation are not simply due to this:
For example, ``Acc. Ens./Flip-Crop81-TTA'' for ResNet-14 (Table~\ref{tab:cifar10_snapshot}) has lower variability across all metrics than ``All Nondeterminism Sources'' for ResNet-18 (Table~\ref{tab:generalization}), despite the ResNet-18  having higher accuracy (92.0\% vs 94.9\%).
Similar trends hold for ResNet-10 with Acc. Ens. and TTA (Table~\ref{tab:linear_onehidden}) vs ResNet-14 (88.7\% vs 89.8\%) and ResNet-6 vs ResNet-10 (77.9\% vs 86.1\%), albeit only visible on certain variability metrics in the latter case.

\end{document}